\documentclass[10pt,twocolumn,letterpaper]{article}

\usepackage[pagenumbers]{wacv} 

\usepackage{graphicx}
\usepackage{amsmath}
\usepackage{amssymb}
\usepackage{booktabs}
\usepackage{multirow}
\usepackage{algorithm}
\usepackage{algpseudocode}
\usepackage{pifont}

\usepackage[pagebackref,breaklinks,colorlinks]{hyperref}
\usepackage{subcaption}

\usepackage[capitalize]{cleveref}
\crefname{section}{Sec.}{Secs.}
\Crefname{section}{Section}{Sections}
\Crefname{table}{Table}{Tables}
\crefname{table}{Tab.}{Tabs.}

\NewDocumentCommand{\jeongh}
{ mO{} }{\textcolor{red}{\textsuperscript{\textit{Jeonghwan}}\textsf{\textbf{\small[#1]}}}}

\NewDocumentCommand{\man}
{ mO{} }{\textcolor{orange}{\textsuperscript{\textit{mankeerat}}\textsf{\textbf{\small[#1]}}}}

\NewDocumentCommand{\revanth}
{ mO{} }{\textcolor{green}{\textsuperscript{\textit{Revanth}}\textsf{\textbf{\small[#1]}}}}

\def\wacvPaperID{2421} 
\def\confName{WACV}
\def\confYear{2025}
\begin{document}
\title{Search and Detect: Training-Free Long Tail Object Detection via Web-Image Retrieval}


\author{
Mankeerat Sidhu\textsuperscript{1}, 
Hetarth Chopra\textsuperscript{1}, 
Ansel Blume\textsuperscript{1}, 
Jeonghwan Kim\textsuperscript{1}, 
Revanth Gangi Reddy\textsuperscript{1}, 
Heng Ji\textsuperscript{1}\\
\textsuperscript{1}University of Illinois Urbana Champaign, Urbana, USA\\
{\tt\small \{mssidhu2, hetarth2, blume5, jk100, revanth3, hengji\}@illinois.edu}
}
\maketitle

\begin{abstract}
    In this paper, we introduce SearchDet, a training-free long-tail object detection framework that significantly enhances open-vocabulary object detection performance. SearchDet retrieves a set of positive and negative images of an object to ground, embeds these images, and computes an input image--weighted query which is used to detect the desired concept in the image. Our proposed method is simple and training-free, yet achieves over 48.7\% mAP improvement on ODinW and 59.1\% mAP improvement on LVIS compared to state-of-the-art models such as GroundingDINO. We further show that our approach of basing object detection on a set of Web-retrieved exemplars is stable with respect to variations in the exemplars, suggesting a path towards eliminating costly data annotation and training procedures.
\end{abstract}

\section{Introduction}
\label{sec:intro}
The proliferation of the web as a repository of image-text data has drastically improved access to data used to train neural object detection models. Modern deep-learning models rely heavily on this colossal cache of data to train and improve their representations. Such datasets are often used to pre-train vision-language models such as CLIP \cite{radford2021learning}, GLIP \cite{li2022grounded}, GroundingDINO \cite{liu2023grounding}, and T-Rex2 \cite{jiang2024trex}. While models such as GroundingDINO and GLIP have achieved substantial advances in zero-shot object detection by reducing the pre-train-to-downstream task discrepancy, further improving the performance of these models necessitates either continual pre-training or additional task-specific finetuning that incurs extra costs.
    
    


\begin{figure}[h]
    
    \begin{subfigure}[b]{0.5\textwidth}
        \centering
        \includegraphics[width=\textwidth]{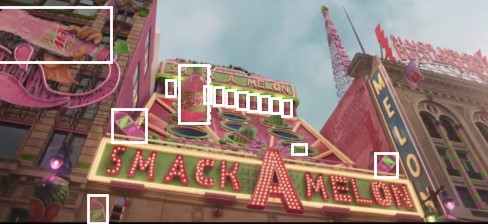}
        \caption{Mountain Dew}
        \label{fig:top_image}
    \end{subfigure}
    

    
    \begin{subfigure}[b]{0.23\textwidth}
        \centering
        \includegraphics[width=\textwidth]{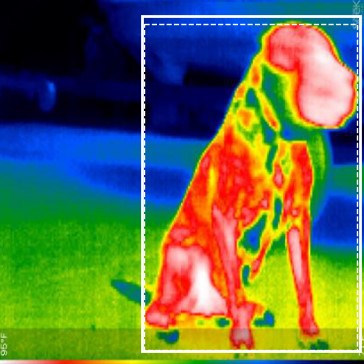}
        \caption{Dog}
        \label{fig:dog1_image}
    \end{subfigure}
    \hspace{0.14cm} 
    \begin{subfigure}[b]{0.23\textwidth}
        \centering
        \includegraphics[width=\textwidth]{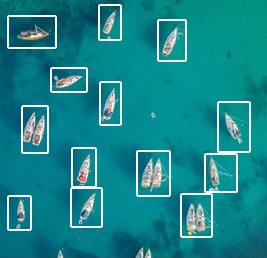}
        \caption{Aerial Boat}
        \label{fig:dog2_image}
    \end{subfigure}

    \caption{Detection results for label ``Mountain Dew''. While the GroundingDINO, one of the state-of-the-art zero-shot object detection methods, fails to capture the Mountain Dew bottles in the image displayed in the figure, SearchDet manages to ground every instance of Mountain Dew that appears in the image. Detection results for other classes ``Dog" and ``Aerial Boat".}
    \label{fig:two_side_images}
\end{figure}

A crucial aspect of the aforementioned models' training involves data indexed by search engines like Google, which provide easy access to high-recall sets of images for a text query. 
This capability opens up several intriguing possibilities for grounding using search engines. 
First, search engines can retrieve relevant images for a given text label, enhancing the specificity of grounding. Secondly, unlike traditional neural models that encode information into their parameters, search engines operate with a retrieval-based framework, utilizing databases that serve as a continuously expanding object ``memory." Integrating such external memory, i.e., the web database of images, during inference provides a potential path to obviate the need for additional training while improving performance on object detection.


In this paper, we propose \textit{SearchDet} (Search and Detect; pronounced ``searched it''), an inference-stage, training-free, long-tail object detection framework that drastically improves performance while avoiding additional finetuning. By leveraging web-retrieved positive and negative image pairs, we calculate attention scores against the query image to perform a weighted averaging of the positive and negative images (\S \ref{method:attention_pool}), ensuring that the query embedding is refined for accurate object detection. The query image representation is generated using DINOv2 \cite{oquab2023dinov2} and compared against the web-retrieved image sets using similarity scores. The process is then followed by frequency-based adaptive thresholding to dynamically determine which regions in the image most likely contain the object by using a binning technique. Finally, by combining information from SAM region proposals and similarity heatmaps, our method generates more precise and robust object boundaries. To summarize, our contributions are as follows:


\begin{itemize}
    \item An inference stage framework for open-vocabulary object detection significantly improves on the state-of-the-art, achieving 48.7\% mAP improvement on ODinW and 59.1\% mAP improvement on LVIS. Our framework avoids the need for additional finetuning or continual pre-training to enhance object detection performance by simply searching for web images and comparing them to input image regions.
    \item Our proposed framework demonstrates the effectiveness and potential of using the web as an external, dynamic memory that provide a stable set of support images for inference stage representations.
\end{itemize}

\section{Related Work}
\label{sec:related}

\subsection{Open-Vocabulary Object Detection}
Open-vocabulary object detection involves grounding text labels of objects within an image using bounding boxes. Open-vocabulary object detection differs from standard object detection in that any text label may be provided, whereas standard object detection models assume a set of fixed object classes to detect.

Existing open-vocabulary object detection models such as GroundingDINO \cite{liu2023grounding}, T-Rex2 \cite{jiang2024trex}, OWL-ViT \cite{minderer2022simple}, and GLIP \cite{li2022grounded} utilize advances in vision-language research to extend closed-vocabulary object detection to an open-vocabulary setting. They frequently utilize a form of contrastive language-image pre-training \cite{radford2021clip}, which pairs text captions with object regions embedded with pre-trained text and image models. While this allows for fast inference times and the flexibility of an open vocabulary, this kind of training is time-consuming and resource-intensive, requiring large amounts of paired image-text data to achieve strong object detection performance. 

By contrast, our method utilizes existing tools and requires \textit{no additional training}. Specifically, we use pre-trained segmentation \cite{kirillov2023sam,ke2024hqsam,ravi2024sam2} and backbone models \cite{oquab2023dinov2, liu2021swin} with web-based retrieval to achieve strong open vocabulary performance. The search engine used for image retrieval serves as an ever-expanding, perpetually improving link between the text and image modalities, a bottleneck that most open vocabulary detection models can address only through further pre-training on larger amounts of data.

\subsection{Few-shot Object Detection}
Few-shot object detection (FSOD) \cite{antonelli2022fsod_survey} is the task of detecting an object in an image when provided a few ``support'' examples of a class. Such methods use techniques from the broader few-shot learning literature including using class prototypes \cite{kang2019fewshotdetectionreweighting,zhang2023detecteverything,snell2017prototypicalnetworks} to represent the class to detect, and meta-learning \cite{zhang2021metadetr,wu2020metalearningfsod, finn2017maml} to train networks to adapt their parameters and representations from few examples. In our work we utilize a set of support images to generate a ``query embedding'', which can be regarded as a class prototype, to detect objects. Like prior FSOD works, our query embedding can be generated from as few as one support image. Unlike most prior works, however, our method involves no training of an object detection system and relies entirely on representations derived from a frozen image backbone.

\begin{figure*}[htbp]
    \centering
    \includegraphics[width=1\textwidth]{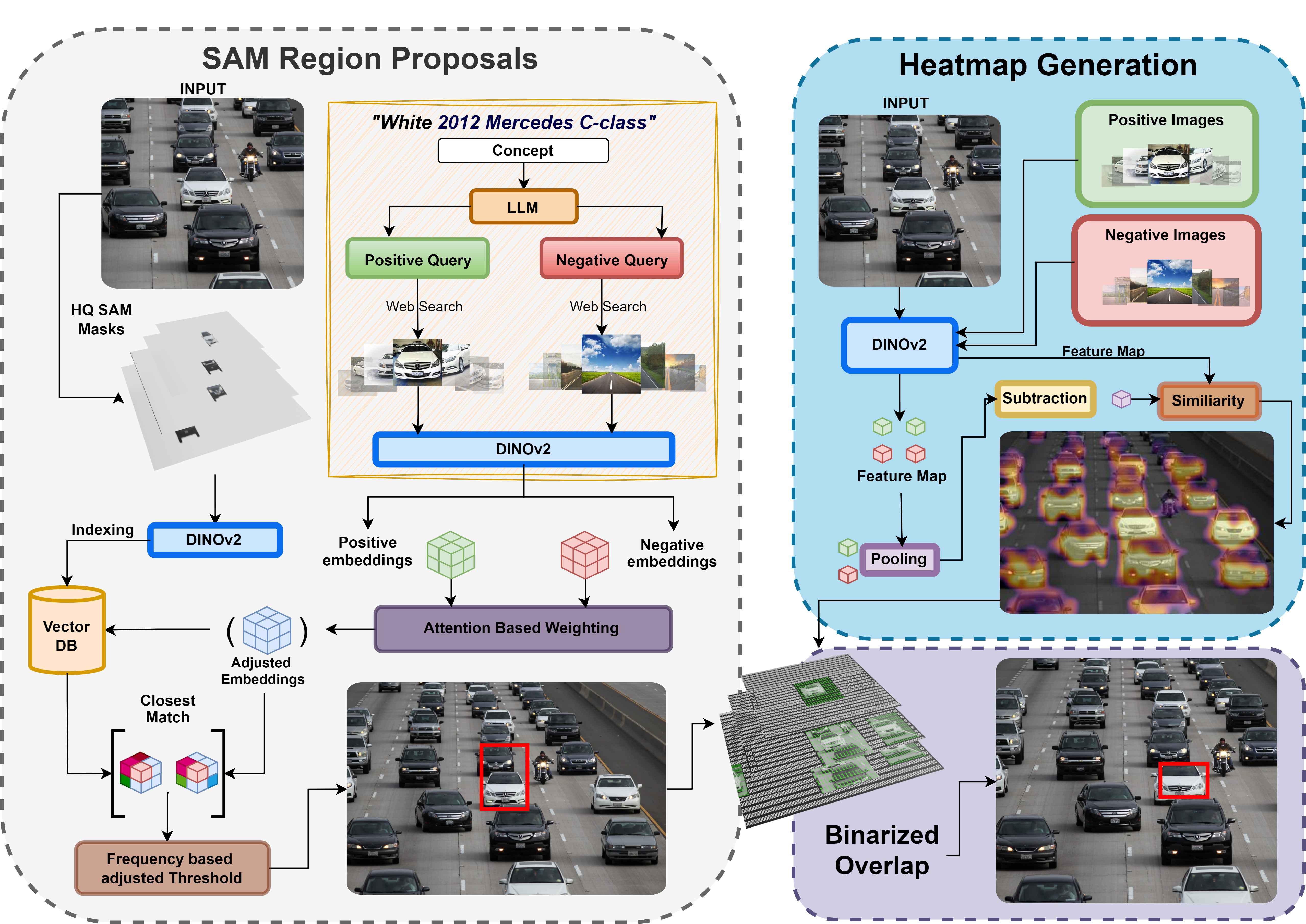}
      \caption{ The entire architecture of our method. We compare the adjusted embeddings, produced by the DINOv2 model, of the positive and negative support images, with the relevant masks extracted using the SAM model to provide an initial estimate of our segmentation BBox. We again use DINOv2 for generating pixel-precise heatmaps which provide another estimate for the segmentation. We combine both these estimates using a binarized overlap to get the final segmentation mask.}
    \label{fig:architecture}
\end{figure*}

\subsection{Image Segmentation}
Image segmentation is the task of grouping pixels of an input image into semantically coherent ``segmented'' parts. This differs from semantic segmentation, which requires classifying each image pixel into a set of predefined classes. The Segment Anything Model (SAM) \cite{kirillov2023sam, ravi2024sam2} and its variants \cite{ke2024hqsam, li2023semantic_sam} are the state-of-the-art in image segmentation, providing highly accurate segmented entities for an input image. 

In our work, we utilize the image regions output by HQ-SAM to generate region proposals for object grounding locations. This is akin to classical works in object detection which generate bounding box proposals before classifying and further refining their position \cite{girshick2015fast_rcnn, ren2016faster_rcnn}. We perform no additional finetuning on HQ-SAM and use its output masks as proposals out of the box.



\section{SearchDet: Object Detection on Web Images}
SearchDet is our proposed object detection framework designed to ground objects using web-retrieved images accurately. For a given \texttt{(image, object label)} pair, our method starts by retrieving positive web images corresponding to the object label, along with negative images to exclude from the image representation. We use these retrieved images to independently generate two object ``queries'' with an attention mechanism, one which detects the image's SAM regions containing the object and one which localizes the object with a similarity heatmap. An adaptive thresholding technique filters the SAM regions to extract those that most closely match the query. Finally, we take the intersection of the most highly ranked SAM regions and our similarity heatmap to output the final predictions. \autoref{fig:architecture} outlines our approach.

In this section, we first describe our approach of web-based image retrieval. We discuss the importance of how we retrieve positive and negative images, our attention-based query-generation method, how we utilize SAM region proposals to enhance detection precision, and frequency-based thresholding to output a set of object regions to improve recall.

\subsection{Web Retrieval of Exemplars}

In this section, we detail the process of retrieving both positive and negative exemplars from the web for concept grounding, and how these exemplars are processed to improve detection accuracy. The structure of the following subsections is designed to walk through the necessity of using negative examples to isolate target objects, followed by an explanation of the attention-based query adjustment, and finally, the adaptive thresholding applied to filter the retrieved masks. This structured approach highlights how each component contributes to precise object localization. To detect the object label in the input image, we start by retrieving image exemplars of the object from the web. We utilize a search engine (in our case, Google) to download a set of images that represent the object of interest, both for the positive and negative query. The top five images from the search results are selected; without any other pre-processing and are passed to DINO-V2 for the attention-based embedding weighting method described in Section \ref{method:attention_pool}.

\subsubsection{Necessity of Negatives}
\label{sec:necessity_of_negatives}
\begin{figure}[t!]
    \centering
    \begin{subfigure}[b]{0.5\textwidth}
        \centering
        \includegraphics[width=0.75\textwidth, trim={0 0 0 2cm}, clip]{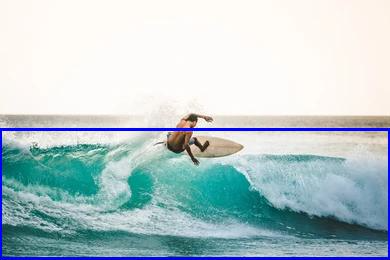}
        \caption{Without including negative support image samples}
        \label{fig:top_image}
    \end{subfigure}
    
    \vspace{0.5cm}
    
    \begin{subfigure}[b]{0.5\textwidth}
        \centering
        \includegraphics[width=0.75\textwidth, trim={0 0 0 2cm}, clip]{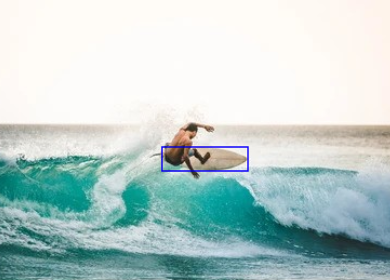}
        \caption{After including negative support image samples}
        \label{fig:bottom_image}
    \end{subfigure}
    \caption{Illustration of our method providing more fine-grained masks after including the negative support images. The negative query (here waves) helps our method, in a way, to not accidentally relevant areas, and only focus on areas represented by the positive query (here surfboard).  }
    \label{fig:positive_vs_negative}
\end{figure}

Retrieved images contain associated objects that are present across all images, making it challenging to isolate the object of interest. For example, many images of a surfboard retrieved from the web also contain water or waves, making it difficult to tightly localize the surfboard. We utilize ``negative queries'' to isolate the target object from these common associations. For a given object label (the positive search query), we utilize a large language model, Microsoft Phi-3-mini-4k-instruct\cite{abdin2024phi}, to generate negative queries using in-context examples, such as ``waves'' for a surfboard or ``food'' for a fork. These negative queries represent opposing or confounding concepts that are likely to interfere with object detection when matching vector representations. Subtracting query embeddings generated from these negative images from the positive embeddings helps isolate the object representation. An example of detection with and without negative queries is shown in \autoref{fig:positive_vs_negative}.

\subsubsection{Attention-based Query Generation}
\label{method:attention_pool}
\autoref{sec:necessity_of_negatives} described how we utilize a set of ``negative images'' to subtract out undesirable negative objects. To assist with the extraction of the most pertinent features, we utilize an attention-based approach taking into account the query image to pool the positive and negative image embeddings. This method leverages a weighted combination of positive and negative embeddings, enhancing the matching areas of the image and the positive embedding while diminishing the matching areas of the image and the negative embedding. This approach is in contrast to directly subtracting the mean-pooled negative embedding from the mean-pooled positive embedding, which can result in over-correction and loss of essential information.

Denote the query embedding by \( \mathbf{q} \in \mathbb{R}^d \) (e.g., \textit{surfboard with waves}), the set of positive embeddings as \( \mathbf{E}_{\text{pos}} = \{\mathbf{e}_{\text{pos}, 1}, \dots, \mathbf{e}_{\text{pos}, n_{\text{pos}}}\} \subset \mathbb{R}^d \), and the set of negative embeddings as \( \mathbf{E}_{\text{neg}} = \{\mathbf{e}_{\text{neg}, 1}, \dots, \mathbf{e}_{\text{neg}, n_{\text{neg}}}\} \subset \mathbb{R}^d \) (e.g., \textit{waves}). Our goal is to generate an adjusted query embedding \( \mathbf{q}_{\text{adjusted}} \) that accentuates the surfboard features while reducing the influence of the waves. We begin by calculating the cosine similarity between the query embedding \( \mathbf{q} \) and each of the positive and negative embeddings:
\begin{equation}
    S_{\text{pos}, i} = \frac{\mathbf{q} \cdot \mathbf{e}_{\text{pos}, i}}{\|\mathbf{q}\| \|\mathbf{e}_{\text{pos}, i}\|}
    ,\, S_{\text{neg}, i} = \frac{\mathbf{q} \cdot \mathbf{e}_{\text{neg}, i}}{\|\mathbf{q}\| \|\mathbf{e}_{\text{neg}, i}\|}
\end{equation}
We then apply a softmax function to the cosine similarities to compute the attention weights for both positive and negative embeddings:
\begin{equation}
    \alpha_{\text{pos}, i} = \frac{e^{S_{\text{pos}, i}}}{\sum_{j=1}^{n_{\text{pos}}} e^{S_{\text{pos}, j}}}, \, \alpha_{\text{neg}, i} = \frac{e^{S_{\text{neg}, i}}}{\sum_{j=1}^{n_{\text{neg}}} e^{S_{\text{neg}, j}}}
\end{equation}
These attention weights ensure that focus is placed on embeddings that are more similar to the query. Using the attention weights, we compute the weighted sums of the positive and negative embeddings:
\begin{equation}
    \mathbf{A}_{\text{pos}} = \sum_{i=1}^{n_{\text{pos}}} \alpha_{\text{pos}, i} \mathbf{e}_{\text{pos}, i},
    \, \mathbf{A}_{\text{neg}} = \sum_{i=1}^{n_{\text{neg}}} \alpha_{\text{neg}, i} \mathbf{e}_{\text{neg}, i}
\end{equation}
The final adjusted query embedding is obtained by subtracting the weighted negative adjustment from the weighted positive adjustment:
\begin{equation}
    \mathbf{q}_{\text{adjusted}} = \mathbf{A}_{\text{pos}} - \mathbf{A}_{\text{neg}}
\end{equation}
By using attention weights, the model selectively emphasizes the most relevant aspects of the positive embeddings while minimizing the influence of shared or irrelevant features present in both positive and negative classes. 

\subsection{SAM Region Proposals}
\label{sec:sam_region_proposals}
To generate region proposals for our object to ground, we turn to the Segment Anything Models (SAM) \cite{kirillov2023sam, ke2024hqsam, ravi2024sam2}. SAM models are trained to take a prompt, most commonly an image coordinate or bounding box, and generate the segmentation for the instance indicated by the prompt. These prompts can be applied in a uniform grid across images, segmenting images into a collection of high-quality object masks. By segmenting images in this way, we obtain region proposals demarcating the most prominent objects in the image. We then check the similarity between each of these region proposals and a query embedding representing the object to determine which regions contain the object to ground. To generate the query embedding, we embed the input image and all support images via DINOv2's \cite{oquab2023dinov2} CLS token. We attention pool (\autoref{method:attention_pool}) the support image embeddings to obtain positive and negative embeddings, then subtract the negative from the positive embeddings to obtain final query embeddings. Next, we generate a region embedding \cite{shlapentokh2024regionbasedrepresentations} for each of the region proposals by masking the input image outside of the region, and then embedding this image using DINOv2's CLS token. We compute the cosine similarity between each region embedding and the query embeddings to obtain similarity scores, then filter these scores via the method of \autoref{method:thresholding} to obtain a set of regions believed to contain the object.

\subsubsection{Frequency-based Automatic Thresholding for Concept Detection}
\label{method:thresholding}

\begin{algorithm}
\caption{Mask Selection and Verification}
\begin{algorithmic}[1]
\Require
  \Statex $Q = \{q_1, q_2, \dots, q_m\}$: set of $m$ adjusted query embeddings
  \Statex $M = \{M_1, M_2, \dots, M_n\}$: set of $n$ segmented masks
  \Statex $d(q_i, M_j)$: Euclidean distance between $q_i$ and $M_j$
  \Statex $T$: predefined acceptance threshold
\Ensure $M_{\text{verified}}$: set of verified selected masks

\Statex \textbf{Definitions:}
\Statex $B = \{b_1, b_2, \dots, b_n\}$: set of bins for distance distribution
\Statex $R = \{ r_{ij} \mid r_{ij} = d(q_i, q_j),\; 1 \leq i < j \leq m \}$: reference distance distribution
\Statex $M(D_{ij}) = j$: mask index function

\State \textbf{Distance Calculation:}
\For{each $q_i \in Q$ and $M_j \in M$}
    \State $D_{ij} \gets d(q_i, M_j)$
\EndFor
\State $D \gets \{ D_{ij} \mid 1 \leq i \leq m,\; 1 \leq j \leq n \}$

\State \textbf{Ordering and Binning:}
\State Sort $D$ into $D_{\text{sorted}} = \{ D_{(1)}, D_{(2)}, \dots, D_{(mn)} \}$ in ascending order
\For{$k = 1$ to $n$}
    \State $b_k \gets \{ D_{(i)} \in D_{\text{sorted}} \mid (k-1)m < i \leq km \}$
\EndFor

\State \textbf{Bin Analysis and Mask Selection:}
\For{each bin $b_k \in B$}
    \For{each mask $M_j$}
        \State $C_k(j) \gets |\{ D_{ij} \in b_k \mid M(D_{ij}) = j \}|$
        \State $P_k(j) \gets C_k(j) / |b_k|$
    \EndFor
    \If{$\max_j P_k(j) > 0.8$}
        \State $M_{\text{selected}}(k) \gets \arg\max_j P_k(j)$
    \Else
        \State $M_{\text{selected}}(k) \gets \text{undefined}$
    \EndIf
\EndFor

\State \textbf{Verification of Selected Masks:}
\State $\mu_R \gets \frac{2}{m(m-1)} \sum_{1 \leq i < j \leq m} r_{ij}$
\State $\sigma_R \gets \sqrt{\frac{2}{m(m-1)} \sum_{1 \leq i < j \leq m} (r_{ij} - \mu_R)^2}$
\For{each $M_j \in M_{\text{selected}}$}
    \State $D_j \gets \{ D_{ij} \mid 1 \leq i \leq m \}$
    \State $\mu_{D_j} \gets \frac{1}{m} \sum_{i=1}^{m} D_{ij}$
    \State $\delta \gets |\mu_{D_j} - \mu_R|$
    \If{$\delta \leq 3\sigma_R$}
        \State $A(M_j) \gets \text{true}$
    \Else
        \State $A(M_j) \gets \text{false}$
    \EndIf
\EndFor
\State \textbf{Output:}
\State $M_{\text{verified}} \gets \{ M_j \in M_{\text{selected}} \mid A(M_j) = \text{true} \}$

\end{algorithmic}
\end{algorithm}

Setting an appropriate threshold for object grounding can be challenging, especially when the concept may or may not be present in the image. Simple thresholding methods, such as those based on fixed percentiles, often fail to adapt to the distribution of scores (such as Euclidean distances between query and masks). These percentile-based methods always output a mask, even when the concept is not in the image, leading to false positives.

To address this limitation, we employ frequency-based adaptive thresholding, which dynamically adjusts the threshold based on the distance distribution. This method adapts to the distribution of distances between adjusted embeddings and segmented masks, providing a robust approach to concept detection and mask selection. 

Let:
\begin{itemize}
    \item $Q = \{q_1, q_2, \dots, q_m\}$ be the set of $m$ adjusted query embeddings (coming from our positive and negative query images).
    \item $M = \{M_1, M_2, \dots, M_n\}$ be the set of $n$ segmented masks, identified by SAM on our target image. 
    \item $d(q_i, M_j)$ be the Euclidean distance between the ith query embedding $q_i$ and the jth segmented mask $M_j$.
    \item $B = \{b_1, b_2, \dots, b_n\}$ be the set of bins for creating distance distribution.
    \item $R = \{r_{ij} \mid r_{ij} = d(q_i, q_j), 1 \leq i < j \leq m\}$ be the set of Euclidean distances between each pair of adjusted embeddings. We define this as the reference distance distribution. 
\end{itemize}


The algorithm aims to identify and verify segmented masks from a target image that closely matches a set of adjusted query embeddings \( Q \). It begins by computing the Euclidean distances \( D_{ij} \) between each adjusted query's embedding \( q_i \) and each mask's embedding calculated by DINOv2 \( M_j \) in the set \( M \). These distances are collected into a set \( D \) and then sorted in ascending order to form \( D_{\text{sorted}} \). The sorted distances are partitioned into \( n \) bins \( B = \{b_1, b_2, \dots, b_n\} \), with each bin containing \( m \) distances (since there are \( m \) queries). For each bin, the algorithm analyzes the distribution of masks by computing the proportion \( P_k(j) \) of distances in bin \( b_k \) that correspond to each mask \( M_j \). If a single mask constitutes more than 80\% of a bin, it is selected as a candidate mask.


In the verification step, the algorithm assesses each selected mask $M_j$ by calculating the mean $\mu_{D_j}$ of all $m$ distances existing in that particular bin as shown in line $29$ in the algorithm. This mean distance reflects the central measure of all distances of the selected mask with adjusted query embeddings. We then calculate the mean of $R$, which is the reference distance distribution, as shown in line $24$ of the algorithm. The distance between these two means is computed as shown in line $29$ of the algorithm. If $\delta$ is greater than 3 standard deviations of the distribution of $R$, where the standard deviation $\sigma_R$ is calculated in step $25$ of the algorithm, we reject the mask. We empirically see that the algorithm effectively filters out less relevant masks, resulting in a robust selection of masks most representative of the adjusted query embeddings.

\subsection{Heatmap Generation}
\label{sec:heatmap_generation}
While the Segment Anything Models typically generate high-quality object regions, they may fail to detect key object regions or may not accurately output objects of interest' boundaries as seen in \autoref{fig:architecture}, where the bounding box generated from a SAM region alone contains two cars. More generally, if we rely only on the SAM regions and the model fails to generate a region corresponding to the object of interest, then we are unable to ground the object accurately. Hence, we adopt a heatmap generation method to ground the object without reliance on preexisting boundaries.

To generate a heatmap of likely object locations, we independently embed the input image and all of the positive and negative images to obtain patch features. We average pool each of these features to obtain a single embedding for each image, then pool the positive and negative image features into a single positive and negative embedding via the same process as \autoref{method:attention_pool}. Subtracting the negative from the positive embedding yields our final query embedding. We then compute the cosine similarity of this query embedding with the input image's upsampled patch features, generating a heatmap of object locations.

\subsection{Joint Object Grounding}
\label{sec:joint_grounding}
We enhance the accuracy of bounding boxes by combining information from both segmentation masks and heatmaps. First, we take the filtered region proposals as the output of \autoref{sec:sam_region_proposals} and the heatmap of \autoref{sec:heatmap_generation}. To refine the object location, we binarize the heatmap by thresholding its brightest regions, setting those areas to 1 and the rest to 0. We then calculate the intersection between each SAM mask and the binarized heatmap. Each region will output a single bounding box, as long as it has a nonempty intersection with the heatmap. If the mask is incomplete or inaccurate, the heatmap can provide complementary information (and vice versa) as illustrated in \autoref{fig:architecture}.

\section{Experiments}


\begin{table*}[htbp]
    \renewcommand{\arraystretch}{1.1} 
    \centering
    \caption{Few-Shot results of our method versus other state-of-the-art methods. We see significant jumps in performance across a diverse set of compiled datasets, especially ODinW and Roboflow-100, which have image annotations that are fine-grained and consist of concepts seen in the wild.}
    \begin{tabular}{l c c c c c}
        \toprule
        \multirow{2}{*}{\textbf{Model}} & \multirow{2}{*}{\textbf{Backbone}} & \multirow{1}{*}{\hspace{10pt}\textbf{COCO}} & \multirow{1}{*}{\hspace{10pt}\textbf{LVIS}} & \multirow{1}{*}{\hspace{10pt}\textbf{ODinW-35}} & \multirow{1}{*}{\hspace{10pt}\textbf{Roboflow100}} \\ 
         &  & \hspace{10pt}val2017 & \hspace{10pt}minival-1203 & \hspace{10pt}val & \hspace{10pt}val \\
         \midrule
        GLIP-L & Swin-L & 49.8 & 26.9 & 23.4 & 8.6\\
        DINOv & Swin-L & 46.2 & - & 15.7 & -\\ 
        GroundingDINO-L & Swin-L & 48.4 & 27.4 & 22.3 & 8.3 \\ 
        T-Rex2 (Text) & Swin-L & 52.2 & \textbf{45.8} & 22.0 & 10.5 \\ 
        T-Rex2 (Visual-G) & Swin-L & 46.5 & 45.3 & 27.8 & 18.5 \\ 
        \textbf{SearchDet (Ours)} & DINOv2-L & \textbf{59.3} & 43.6 & \textbf{33.1} & \textbf{27.9} \\ 
        \bottomrule
    \end{tabular}
\end{table*}

\begin{table}[h!]
\centering
\caption{Performance of various methods on 10-shot. Our method outperforms the current state of the art by 16\%.}
\begin{tabular}{lcc}
\toprule
\multirow{2}{*}{Method} & \multirow{2}{*}{Finetuned on Novel} & \multicolumn{1}{c}{10-shot} \\
\cmidrule(lr){3-3}
& & mAP50 \\
\midrule
FSRW \cite{DBLP:conf/iccv/KangLWYFD19} & \ding{55} & 12.3 \\
Meta R-CNN \cite{DBLP:conf/iccv/YanCXWLL19} & \ding{55} & 19.1 \\
TFA \cite{DBLP:conf/icml/WangH0DY20} & \ding{51} & 19.2 \\
Multi-Relation Det \cite{DBLP:conf/cvpr/Fan0TT20} & \ding{55} & 31.3 \\
FSCE \cite{DBLP:conf/cvpr/SunLCYZ21} & \ding{55} & 30.5 \\
Retentive RCNN \cite{DBLP:conf/cvpr/FanMLS21} & \ding{51} & 19.5 \\
HeteroGraph \cite{DBLP:conf/iccv/HanHHMC21} & \ding{55} & 23.9 \\
Meta Faster RCNN \cite{DBLP:conf/aaai/HanHMHC22} & \ding{51} & 25.7 \\
LVC  \cite{DBLP:conf/cvpr/KaulXZ22} & \ding{51} & 34.1 \\
CrossTransformer \cite{DBLP:conf/cvpr/HanMH0C22} & \ding{55} & 30.2 \\
DiGeo \cite{DBLP:conf/cvpr/MaNXHHC23} & \ding{51} & 18.7 \\
DE-ViT \cite{zhang2024detectexamples} & \ding{55} & 52.9 \\
\midrule
\textbf{SearchDet (Ours)} & \ding{55}  & \textbf{61.4} \\
\bottomrule
\end{tabular}
\end{table}

\subsection{Datasets and Metrics}
We demonstrate the effectiveness of our method on four settings - the COCO detection benchmark \cite{lin2014microsoft}, LVIS \cite{gupta2019lvis}, OdinW \cite{li2021glip}, and Roboflow-100 \cite{2211.13523}. We do not use the training datasets, as ours is a training-free method; instead, we focus on the COCO-2017-val split (80 classes), and the LVIS minival-version-1.0 (1203 classes), while using the full OdinW and Roboflow-100 validation splits. We use the class names as the concept, get the negative query name from an LLM call, and extract 10 positive and negative support images from the web for all datasets. We pass them to our method to get a precise mask which is compared to the ground truth. We compare our method to different state-of-the-art open vocabulary object detection methods including GLIP DINOv \cite{li2024dinov_prompting}, Grounding Dino \cite{liu2023grounding}, and T-Rex2 \cite{jiang2024trex}. DINOv and T-Rex2 are especially pertinent, as they also require in-context images to perform object detection.  We also test our method in a few-shot setting (10-shot) and have a 16.1\% performance increase over the SOTA.

\subsection{Performance of Web Grounding}
SearchDet demonstrates significant performance improvements across multiple benchmark datasets compared to state-of-the-art object detection models. Our experiment uses five support images (both positive and negative) for our given concept, and each image takes approximately 3 seconds to run on a single NVIDIA-V100 GPU, however, the time may vary since we scrape these images from the web.  On the COCO val2017 dataset, SearchDet achieves a score of 59.34, outperforming all compared methods with improvements ranging from 13.68\% (vs. T-Rex2 Text on COCO) to 28.44\% (vs. DINOv2 on COCO). While slightly behind T-Rex2 variants on LVIS, SearchDet still shows substantial gains over other methods, with up to 62.08\% improvement (vs. GLIP-L on LVIS). We posit that our method lags on LVIS because our experiment chooses just 5 support images. This is supported by our stability analysis of \autoref{sec:stability}, where we see that including more images leads to steady growth in the mAP. 

Our method's performance is particularly noteworthy on more diverse and challenging datasets: for ODinW-35, it surpasses all other methods with improvements from 19.32\% (vs. T-Rex2 Visual-G on ODinW-35) to 111.27\% (vs. DinoV on ODinW-35), and on Roboflow100, it achieves remarkable gains of up to 236.27\% (vs. GroundingDINO-L on Roboflow100). These results, especially the consistent and substantial improvements on ODinW-35 and Roboflow100, suggest that SearchDet offers enhanced generalization and robustness across varied object detection tasks, representing a significant advancement in the field. One notable observation from our analysis is that the mAP scores can be significantly improved when the provided label is more descriptive. For instance, in the OdinW dataset, some labels consist of generic terms such as ``20'' or ``boat.'' Searching for ``20'' or ``boat'' often retrieves irrelevant images, despite the annotated images containing more specific concepts like ''20 dollar bill" or ''aerial view of boat."

\begin{figure}[t!]
    \centering
    \begin{subfigure}[b]{0.5\textwidth}
        \centering
        \includegraphics[width=\textwidth]{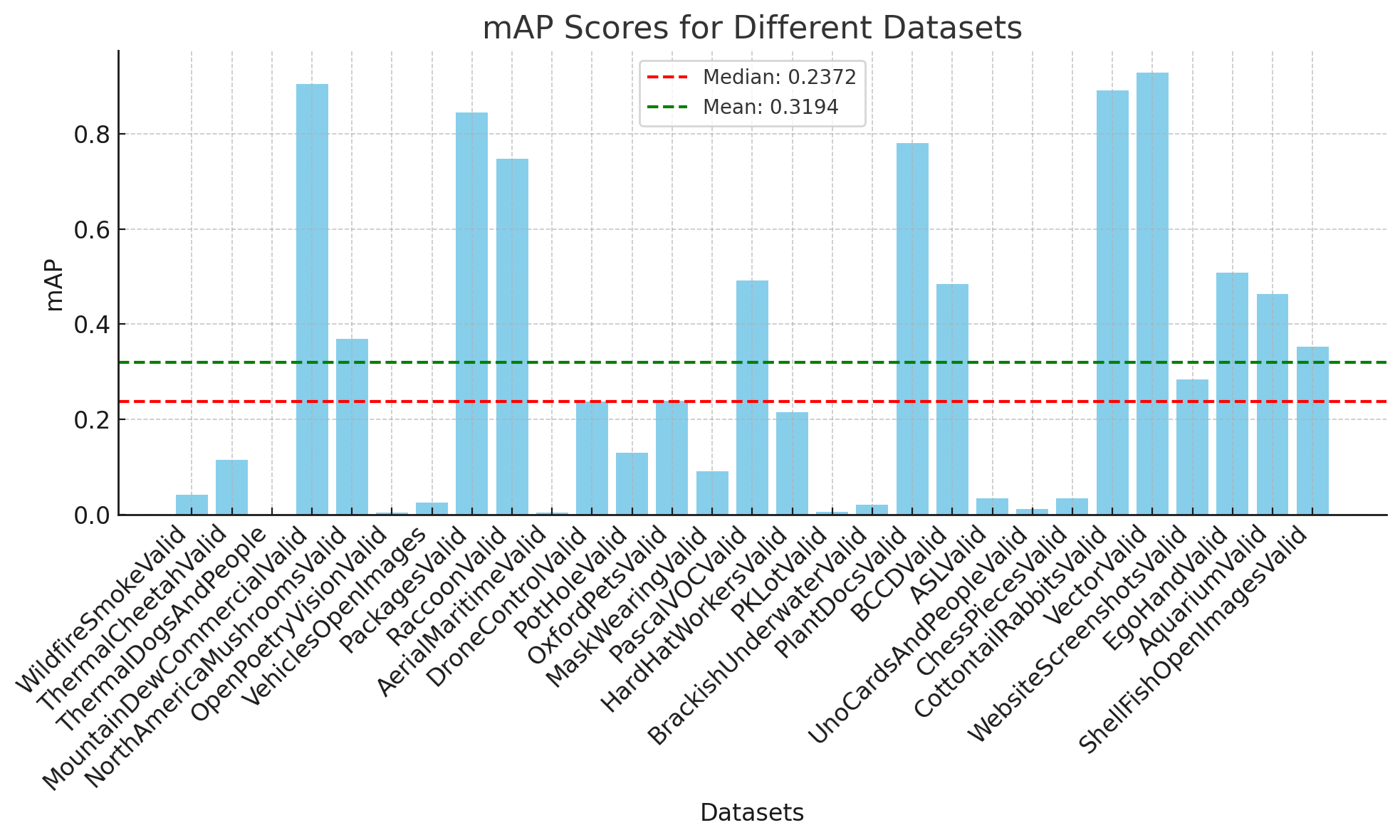}
        \caption{Without passing the name of the folder as the concept}
        \label{fig:top_image}
    \end{subfigure}%
    
    \vspace{0.5cm}
    
    \begin{subfigure}[b]{0.5\textwidth}
        \centering
        \includegraphics[width=\textwidth]{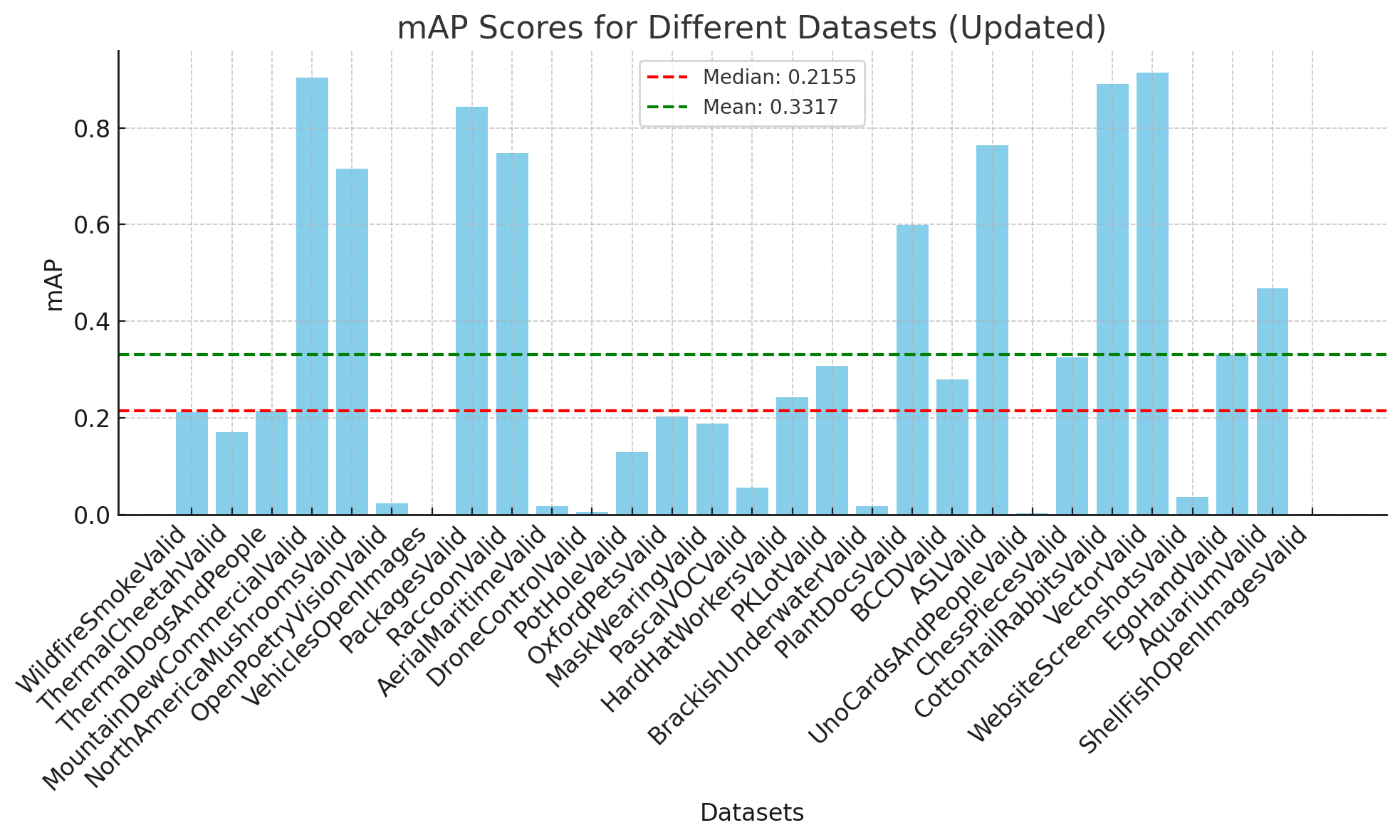}
        \caption{With passing the name of the folder as the concept}
        \label{fig:bottom_image}
    \end{subfigure}

    \caption{A comparison of our method's mAP on the OdinW Dataset under different concept names. We see a 3.85\% increase in the mean mAP just by including the name of the dataset (for example WildfireSmoke) with the name of the concept in the image.}
    \label{fig:odinw_dataset_analysis}
    
\end{figure}


\begin{figure}[ht!]
    \centering
    \includegraphics[width=0.45\textwidth]{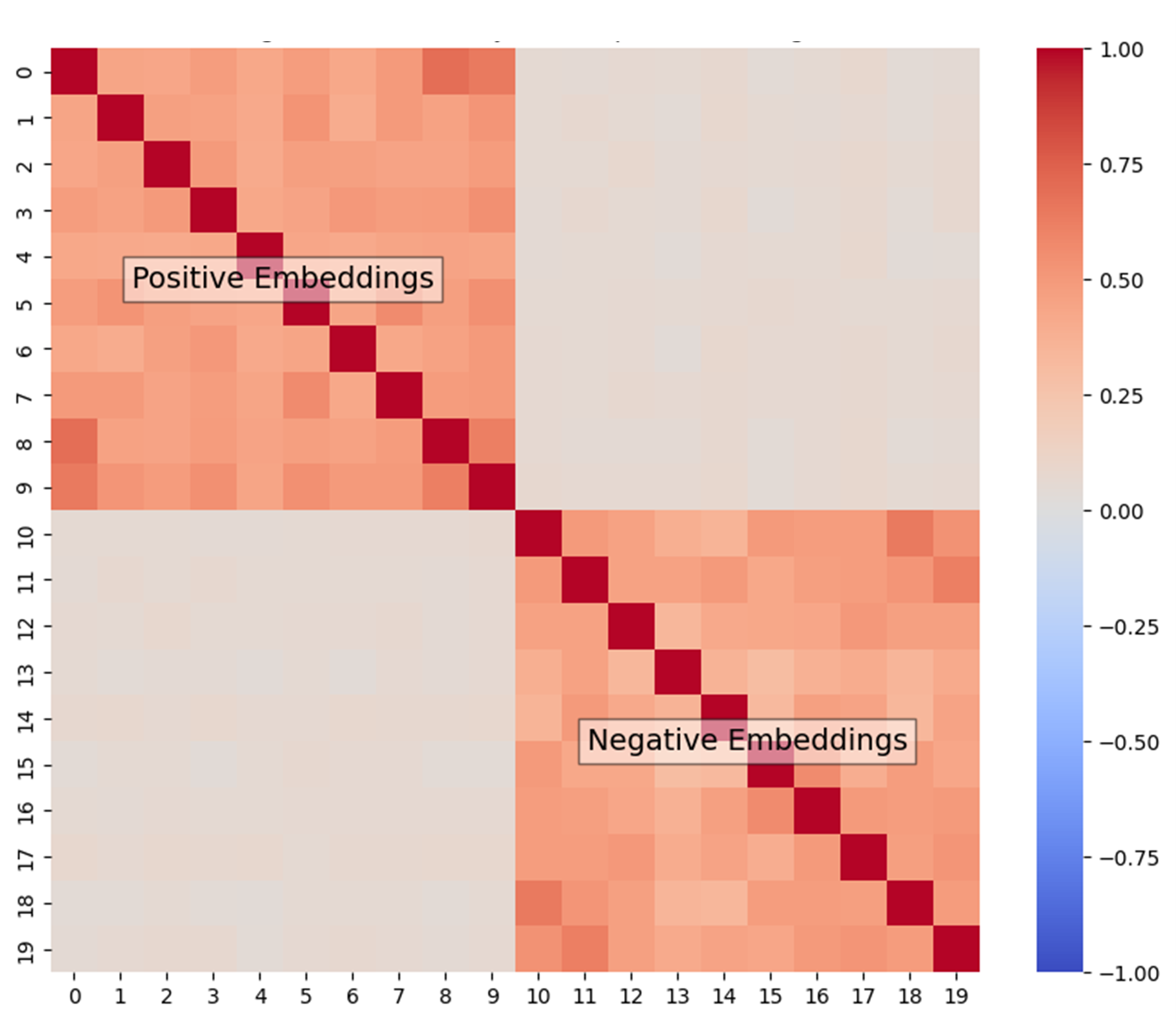}
    \caption{Stability analysis showcasing the cosine similarity of embeddings generated from the positive and negative support images (ten images of each), averaged across all eighty classes in the COCO dataset. The high similarity scores demonstrate the stability of our method, which exhibits consistent patterns in embedding similarities despite the dynamic nature of web-based image retrieval.}
    \label{fig:stability_2}
\end{figure}

\begin{figure}[ht!]
    \centering
    \includegraphics[width=0.48\textwidth]{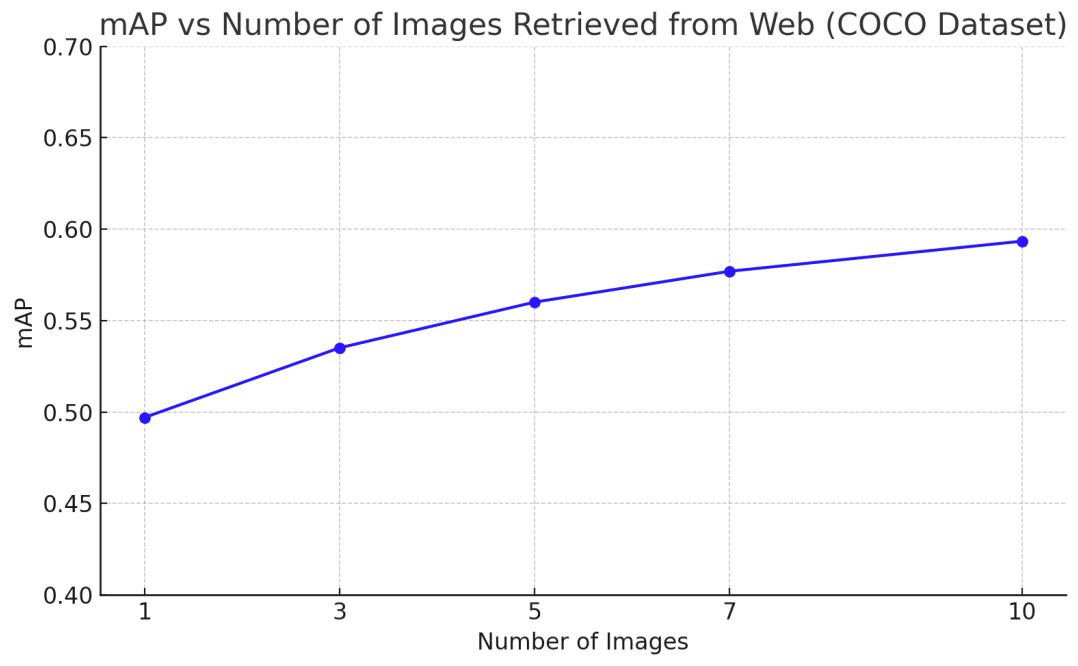}
    \caption{Relationship between the number retrieved web images versus performance on COCO.}
    \label{fig:image_number_ablation}
\end{figure}

\subsection{Stability Analysis}
\label{sec:stability}

Two important considerations are how the number and choice of retrieved support images affect the performance of our method. To evaluate the first, we vary the number of support images provided to our method, observing that even a single positive and negative support image is good enough to generate an mAP of 0.4970 on the COCO Dataset. However, we see a steady increase in mAP as the number of retrieved instances increases, as seen in \autoref{fig:image_number_ablation}. With 10 images, we obtain an mAP of 0.5934, an increase of 19.41\% over a single image. This study indicates that not only does our method's performance scale with the number of retrieved images but also that it is stable enough that we don't lose out on performance with few images. 

To study the similarity between different sets of retrieved images, we inspect the relationship between their embeddings. We use the set of COCO labels as queries and extract 10 positive and negative support images for each class. We then average the cosine similarities across all classes \autoref{fig:stability_2}. We notice that even though the internet is a dynamic space, the embeddings maintain consistent similarities. In particular, we see that all positive support images exhibit strong similarity and the negative support images do the same.  This observation suggests that the LLM-based method to generate a negative class helps to retrieve images that are significantly different from their positive counterparts. We further observed that by downloading images on different days of the week, we would obtain different sets of images. However, these different image sets did not have a significant effect on performance. These results together suggest that the retrieved images are stable enough to not be adversely affected by the dynamic nature of web retrieval.  


    
    


\subsection{Ablations}
\begin{table}[ht]
    \centering
    \caption{Performance comparison on COCO Dataset}
    \begin{tabular}{@{}p{0.65\columnwidth}c@{}}
        \hline
        \textbf{Ablation} & \textbf{mAP} \\
        \hline
        Our Method & 59.34 \\
        Only Positive Support Images & 45.80 \\
        No RoI Refinement with Heatmaps & 51.07 \\
        Mean-Pooling of Support Images & 55.47 \\
        \hline
    \end{tabular}
    \label{tab:ablations}
\end{table}

We discuss different ablations of our method on the COCO Dataset in \autoref{tab:ablations}. First, using only positive support images and removing the negative concept images leads to a significant drop in mAP (approximately 22.82\%). Next, we see the usefulness of refining the SAM object predictions using heatmaps from the 13.94\% decrease in mAP when only using SAM masks. Finally, we see the usefulness of our attention-based pooling by comparing it to mean-pooling of the support images. We find that this results in a 6.5\% decrease in mAP value. These ablations demonstrate the importance of each component in our method.



\section{Conclusion}
In this paper, we presented SearchDet, a training-free inference stage framework that leverages web-retrieved images for long-tail open-vocabulary object detection. Our experiments demonstrate that SearchDet not only outperforms existing state-of-the-art models like GroundingDINO and GLIP-L but also that SearchDet shows robustness against variations in exemplars used for object detection. We see that while the performance of our method is proportional to the number of Web-retrieved images, even a single retrieved image is sufficient for strong performance. Our work opens new avenues for exploration, showing that training-free methods leveraging pre-trained vision models and dynamic web images obtain strong performance without the need to continuously fine-tune or pre-train open-vocabulary detectors.

{\small
\bibliographystyle{ieee_fullname}
\bibliography{egbib}
}

\end{document}